\documentclass[letterpaper,10pt,conference]{ieeeconf}

\IEEEoverridecommandlockouts
\usepackage{graphicx}
\usepackage{subcaption}
\usepackage{amsmath,amssymb}
\usepackage{booktabs}
\usepackage{multirow}
\usepackage{array}
\usepackage{url}
\usepackage{xcolor}
\usepackage[utf8]{inputenc}
\usepackage{censor}
\usepackage{balance}
\usepackage{placeins}
\DeclareUnicodeCharacter{0306}{}

\newif\ifanonymize
\anonymizefalse
\DeclareRobustCommand{\redact}[1]{\ifanonymize\censor{#1}\else #1\fi}

\newcommand{\orin}{Jetson Orin}
\newcommand{\personminkunet}{Person-MinkUNet}

\newcommand{\noid}{\mathrm{NA}}

\title{\LARGE\bf
Beyond Leaderboard Scores: A Deployment-Focused Protocol for Interpretable Tracking Evaluation in Pedestrian-Centric Environments
}

\author{%
\redact{Dominik Wojcikiewicz} and \redact{Diego Paez-Granados}%
\thanks{\scriptsize
\ifanonymize
\censor{This work was supported by the JST Moonshot R\&D (Grant JPMJMS2034-18).}\newline
\censor{All authors are with the Spinal Cord Injury Artificial Intelligence Lab (SCAI),}\newline
\censor{D-HEST, ETH Zurich, Switzerland, and with Digital Health Care and Rehabilitation,}\newline
\censor{Swiss Paraplegic Research (SPF), Switzerland.}%
\else
This work was supported by the JST Moonshot R\&D (Grant JPMJMS2034-18).\newline
All authors are with the Spinal Cord Injury Artificial Intelligence Lab (SCAI), D-HEST, ETH Zurich, Switzerland, and with Digital Health Care and Rehabilitation, Swiss Paraplegic Research (SPF), Switzerland.\newline
(e-mail: dominik.wojcikiewicz@hest.ethz.ch, diego.paez@hest.ethz.ch).%
\fi
}%
}

\begin{document}
\bstctlcite{BSTcontrol}
\maketitle
\thispagestyle{empty}
\pagestyle{empty}

\begin{abstract}

Mobile robots operating among pedestrians need trajectories that become available quickly, remain spatially credible through missed observations, preserve identity, and fit within an embedded computing budget. Aggregate tracking scores provide limited insight into when and how trajectories fail, while varying detector inputs can confound tracker and detector quality. We present a deployment-focused, tracker-only evaluation protocol that uses shared detections to isolate tracker behavior and directly evaluates initialization, detector-gap continuation, identity recovery, close-neighbor association, and load-dependent tracker-step runtime, while Higher Order Tracking Accuracy (HOTA) is retained as a complementary aggregate measure. We apply the protocol to the JackRabbot Dataset and Benchmark (JRDB) using six open-source trackers and our lightweight Pedestrian Reference Tracker (PedRefTrack), together with a GT-assisted variant that estimates the remaining tracker-side gap under idealized association and motion. Under fixed detections, the non-GT trackers span only 24.26\%--29.67\% HOTA yet exhibit markedly different capability profiles. After 1.0\,s without detector support, no tracker without GT assistance maintains spatially correct, same-identity output in more than half of eligible cases, making missing-observation continuation the dominant limitation among the tested properties. Close-neighbor failures are smaller and increase mainly at the shortest separations. Tracker-step runtime on an NVIDIA Jetson Orin is heavy-tailed and load-sensitive, causing several trackers to fall below the 10\,Hz real-time target in crowded frames. The protocol provides a reproducible way to characterize tracker behavior and deployment suitability in pedestrian-centric environments. Code and evaluation scripts are released at \redact{\url{https://github.com/SCAI-Lab/tracker_eval}}.
\end{abstract}

\noindent\textbf{Index Terms---}3D multi-object tracking, pedestrian tracking, tracking evaluation, mobile robotics, LiDAR, HOTA.

\section{Introduction}\label{sec:introduction}
Robots operating in pedestrian-centric spaces such as hospitals, airports, and transit hubs need more than isolated pedestrian positions. A robot path planner uses recent trajectories to estimate walking direction and speed, predict near-term motion, assess collision risk, and choose socially compliant paths \cite{kruse_human-aware_2013,singamaneni_survey_2024,alyassi_social_2025}. Tracking output must therefore arrive with limited delay, preserve identity, remain credible through missed observations, and be computed within the robot's control cycle. Whether a tracker is suitable for deployment depends on its temporal behavior and failure conditions, not only on aggregate tracking accuracy.


Most 3D multi-object tracking (MOT) systems are developed and evaluated on autonomous-driving datasets centered on roadway geometry and vehicle-mounted sensing \cite{zhang_3d_2023,luo_multiple_2021}. Pedestrian-centric navigation presents a different operating regime in which people are observed at close range, nonlinear walking motion is common, and frequent partial occlusions arise from the crowd itself \cite{martin-martin_jrdb_2023,shenoi_jrmot_2020,bae_sit_2023}. Performance established in the driving regime therefore does not by itself characterize the challenges faced by a mobile robot moving among pedestrians. Moreover, aggregate MOT scores summarize performance over an entire test set, providing limited insight into how tracker behavior relates to sequence complexity and deployment-relevant situations \cite{pedersen_motcom_2022,francis_principles_2025}. Attributing these differences to tracker design is further complicated when evaluation is performed with different input detections, since scores then reflect both detection and tracking behavior. Shared detections are established in 2D MOT benchmarking \cite{milan_mot16_2016,dendorfer_motchallenge_2021}, but their use is generally not enforced by 3D MOT and robot-centric pedestrian-tracking benchmarks \cite{geiger_vision_2013,caesar_nuscenes_2020,sun_scalability_2020,martin-martin_jrdb_2023}.

\begin{figure*}[!htb]
\centering
\includegraphics[width=0.92\textwidth]{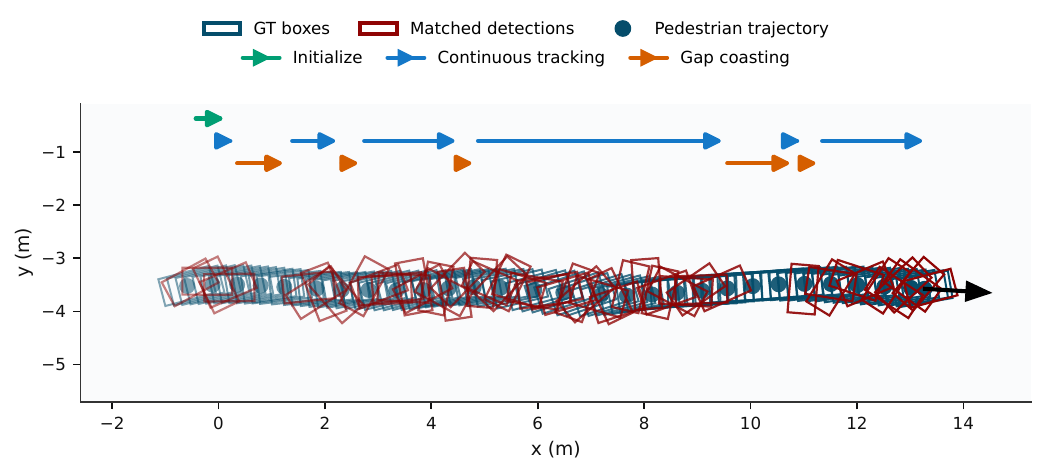}
\caption{\textbf{Trajectory partition for deployment-focused tests.} An example pedestrian trajectory is divided into three complementary regions. Green marks the first $0.5$\,s after detector support begins, blue marks mostly continuous supported motion containing only brief detection gaps, and orange marks longer gaps bounded by supported detections. Together, these regions span the trajectory and define separate event sets for the tests in Sec.~\ref{sec:protocol}.}
\label{fig:protocol_overview}
\end{figure*}

We therefore propose a deployment-focused, tracker-only protocol that evaluates tracker behavior under shared detections (Fig.~\ref{fig:protocol_overview}) with embedded runtime across input loads. Eligible events are defined from GT and detection information so every tracker is evaluated under identical conditions.

Rather than introducing a new composite tracking score, the protocol complements HOTA with directly interpretable success profiles for initialization, continuation through detector gaps, post-gap identity recovery, and close-neighbor association, while separately profiling tracker-step runtime as a function of input load on embedded hardware. As part of the protocol, we implement PedRefTrack, a lightweight pedestrian-oriented reference tracker, together with a GT-assisted variant that uses GT identity and displacement to idealize association and motion estimation, exposing the remaining performance gap associated with these factors. We instantiate the protocol on JRDB \cite{martin-martin_jrdb_2023} to provide an evaluation of how established tracking methods handle deployment-relevant challenges in a pedestrian-centric robot setting. A supplementary video provides an overview of the protocol and case-study results.

\section{Related Work}\label{sec:related}
\subsection{Tracking requirements and evaluation}
Prior work motivates evaluating tracking beyond aggregate accuracy from two complementary perspectives. MOT reviews identify frequent occlusions, track initialization and termination, identity maintenance, and interactions among multiple objects as persistent challenges \cite{luo_multiple_2021}. From the deployment perspective, recent social-navigation evaluation guidelines emphasize scenario-aware benchmarks and metrics tied to the conditions and requirements of the target domain \cite{francis_principles_2025}. Together, these observations motivate evaluating not only overall tracking accuracy but also the temporal conditions under which trajectories remain usable.

These deployment-relevant temporal conditions are only indirectly reflected in standard aggregate MOT metrics. MOTA combines false negatives, false positives, and identity switches, but can be dominated by detection errors and represents association primarily through identity switches \cite{bernardin_evaluating_2008,luiten_hota_2021}. IDF1 instead emphasizes long-term identity consistency through a global one-to-one matching of predicted and GT trajectories, strongly penalizing fragmentation while ignoring association quality outside the selected trajectory pairings \cite{luiten_hota_2021,ristani_performance_2016}. HOTA balances detection and association accuracy while integrating localization across thresholds \cite{luiten_hota_2021}.

\subsection{3D MOT families and representative open-source trackers}
MOT is commonly implemented in a tracking-by-detection paradigm, in which per-frame hypotheses are linked into trajectories through association and lifecycle management \cite{luo_multiple_2021}. Joint detection--tracking formulations integrate association into the detection model, but tracking-by-detection remains the dominant interface among open-source 3D MOT systems \cite{zhang_3d_2023}. Our study therefore focuses on \emph{online}, detection-based methods that consume per-frame 3D boxes.

Representative design families include:
(i) \emph{motion-model + assignment} trackers that combine simple dynamics with cost-based matching, such as AB3DMOT \cite{weng_3d_2020} or the tracking component of CenterPoint \cite{yin_center-based_2021};
(ii) \emph{confidence-aware lifecycle} methods that explicitly handle scores, confirmation, and termination to improve stability under noisy detections, such as SimpleTrack \cite{pang_simpletrack_2022} or CBMOT \cite{benbarka_score_2021};
(iii) \emph{polyhedral/geometry-aware association} frameworks that incorporate richer 3D geometric constraints into matching, including Poly-MOT \cite{li_poly-mot_2023} and Fast-Poly \cite{li_fast-poly_2024};
(iv) \emph{probabilistic tracking} approaches that model track existence and association uncertainty explicitly, including GNN-PMB \cite{liu_gnn-pmb_2023} and FGO-PMB \cite{jin_fgo-pmb_2026};
and (v) \emph{learned association} methods that learn matching costs or association policies from data while retaining a detection-based interface, such as 3DMOTFormer \cite{ding_3dmotformer_2023} and PolarMOT \cite{kim_polarmot_2022}. These families differ primarily in motion prediction, association, and lifecycle management, motivating comparison under common detector inputs.

\subsection{Datasets across tracking domains}
Autonomous-driving benchmarks such as KITTI \cite{geiger_vision_2013}, nuScenes \cite{caesar_nuscenes_2020}, and Waymo \cite{sun_scalability_2020} dominate 3D MOT evaluation but primarily represent vehicle-centric road traffic. Robot-centric pedestrian navigation instead places the ego platform in shared human spaces, where close encounters and dense interactions are more common. SiT explicitly reports substantially more close-proximity robot--pedestrian encounters than several autonomous-driving datasets \cite{bae_sit_2023}.

Robot-centric datasets themselves span diverse operating regimes. JRDB captures natural indoor and outdoor pedestrian activity on a university campus \cite{martin-martin_jrdb_2023}, SiT covers urban scenes in Seoul, including streets, hallways, and crosswalks \cite{bae_sit_2023}, TH\"OR focuses on controlled indoor human--robot scenarios \cite{rudenko_thor_2020}, SCAND records socially compliant navigation demonstrations across varied indoor and outdoor environments \cite{karnan_socially_2022}, and CrowdBot v2 emphasizes dense outdoor crowd navigation \cite{wojcikiewicz_crowdbot_v2_2025}. These datasets therefore expose complementary pedestrian interaction regimes.

\section{Deployment-Focused Tracker-Only Protocol}\label{sec:protocol}
\subsection{Common inputs and matching}
Before evaluating trackers, we define detector support from the shared detections. In each frame, GT boxes and detections are matched one-to-one using Hungarian assignment on XY-center distance, with a $0.30$\,m gate. Tracker outputs are evaluated separately: in each frame, they are assigned one-to-one to GT by maximizing oriented 3D IoU, with matches requiring IoU $\geq 0.30$. This requires meaningful overlap while tolerating some pedestrian-box errors. For tracker $i$, GT trajectory $g$, and time $t$, let $J_i(g,t)$ denote the identity of the accepted tracker output, and set $J_i(g,t)=\noid$ if no output meets the overlap criterion.

Event eligibility, detector-gap duration, and nearest-neighbor difficulty are derived only from GT and the shared detections. Tracker output determines event success, but event eligibility is fixed independently of the tracker. All methods are therefore evaluated on the same events.

\subsection{HOTA and PedRefTrack}
At a localization threshold $\alpha$, HOTA is the geometric mean of detection and association accuracy \cite{luiten_hota_2021}:
\begin{equation}
 \mathrm{HOTA}_{\alpha}=\sqrt{\mathrm{DetA}_{\alpha}\,\mathrm{AssA}_{\alpha}}.
 \label{eq:hota}
\end{equation}
The reported HOTA averages this score over localization thresholds. DetA separates further into detection recall and precision (DetRe and DetPr), AssA into association recall and precision (AssRe and AssPr), and LocA reports spatial alignment. The decomposition provides a split-wide summary of broad detection, association, and localization errors.

To estimate how much performance can be recovered through tracker design, we implement \emph{PedRefTrack}, a lightweight pedestrian-oriented tracking-by-detection reference assembled from established practices. It targets real-time LiDAR inference on embedded hardware such as \orin{} even in crowded frames. PedRefTrack combines a low-cost constant-velocity Kalman filter, moving and stationary retention modes, score-aware confirmation, two-pass IoU and distance association, filtered box geometry, and separate lifetimes for published predictions and hidden identity retention. A key distinction is that PedRefTrack does not choose coasting duration from one fixed timeout or the latest detector score. At the onset of a detector gap, it measures recent pre-correction Kalman residuals to estimate how accurately its motion model predicted that trajectory before support disappeared. A shifted exponential maps this prediction error to a configurable interval between $T_{\min}$ and $T_{\max}$, set to $0.5$ and $2.0$\,s in our experiments. Accurate recent prediction therefore yields a longer published coasting horizon. A second limit predicts, under relative constant velocity, when two pedestrian centers first come within $0.5$\,m and caps the published horizon before that encounter. Moving and stationary tracks can remain hidden for $2.5$ and $5.0$\,s, respectively, so identity recovery does not require publishing an uncertain position throughout the gap.

In \emph{GT-assisted} PedRefTrack, GT identity and displacement disambiguate association and motion for true pedestrian tracks. The tracker still uses the shared detections for box geometry and retains the same confirmation, filtering, coasting, and finite-lifetime decisions. Unmatched detections can still create false-positive tracks. This is not perfect tracking, but a diagnostic estimate of the performance that could be recovered if association and pedestrian motion were known more reliably.

\subsection{Trajectory-level success tests}
The four tests below define the eligible trajectory events and the success criterion used for each deployment behavior.

\paragraph{Tracking in a detector gap}
A bounded gap event $e$ contains GT trajectory $g_e$, the last detector-supported time $t_e^{-}$ before the gap, and the first supported time $t_e^{+}$ afterward. We require $0.5$\,s of continuous detector support before $t_e^{-}$ so that each tracker has time to initialize the pedestrian and estimate its motion. Write $j_{i,e}^{-}=J_i(g_e,t_e^{-})$ for the pre-gap tracker identity. For elapsed gap age $a$, let $\mathcal{E}_{\mathrm{gap}}(a)$ contain the common gaps that last at least $a$. Event success and the corresponding success rate are
\begin{align}
s_{i,e}^{\mathrm{gap}}(a)
&=\mathbf{1}\!\left\{
\begin{aligned}
j_{i,e}^{-}&\neq\noid,\\
J_i(g_e,t)&=j_{i,e}^{-}
\quad \forall t\in(t_e^{-},t_e^{-}+a]
\end{aligned}\right\}, \label{eq:gap_event}\\
S_i^{\mathrm{gap}}(a)
&=\frac{1}{|\mathcal{E}_{\mathrm{gap}}(a)|}
\sum_{e\in\mathcal{E}_{\mathrm{gap}}(a)}s_{i,e}^{\mathrm{gap}}(a).
\label{eq:gap_rate}
\end{align}
Success requires the pre-gap identity to remain spatially accepted at every evaluated time point; missing or spatially invalid output therefore fails the event. Internal identity retention without a published valid position is not credited.

\paragraph{Post-gap identity recovery}
The second test uses the same bounded gaps but ignores all output during the missing interval. It checks whether the first output when detector support returns has the same identity as before the gap:
\begin{equation}
s_{i,e}^{\mathrm{rec}}
=\mathbf{1}\!\left\{
j_{i,e}^{-}\neq\noid
\ \wedge\ 
J_i(g_e,t_e^{+})=j_{i,e}^{-}
\right\}.
\label{eq:recovery_event}
\end{equation}
The recovery profile averages Eq.~\eqref{eq:recovery_event} over the common gaps in each duration range and measures identity retention and re-association independently of spatial output during the gap. It is therefore complementary to Eqs.~\eqref{eq:gap_event}--\eqref{eq:gap_rate}, which require valid published output throughout the interruption.

\paragraph{Close-neighbor association}
We form non-overlapping $1.0$\,s windows $W_e$ after a $0.5$\,s initialization interval within GT-continuous trajectory segments. Detector gaps inside a window may be no longer than $0.2$\,s. A window succeeds only when one spatially accepted tracker identity covers every time point:
\begin{equation}
s_{i,e}^{\mathrm{cont}}
=\mathbf{1}\!\left\{
\exists j\neq\noid:\ 
J_i(g_e,t)=j\quad \forall t\in W_e
\right\}.
\label{eq:continuity_event}
\end{equation}
We report one minus the mean of Eq.~\eqref{eq:continuity_event} as the failure rate and stratify it by nearest-neighbor distance. Difficulty is the 10th percentile of the pedestrian's GT nearest-neighbor distance over the window, so a brief close encounter is not hidden by otherwise larger separation. Because windows may contain gaps up to $0.2$\,s, this profile measures close-neighbor association rather than pure frame-to-frame association.

\paragraph{Stable initialization latency}
Let $\tau_e$ be the first detector-supported time of an eligible GT trajectory. For $0\leq a\leq0.5$\,s, an event has been stably initialized by delay $a$ if some start within the first $a$ seconds is followed by $1.0$\,s of uninterrupted, spatially accepted output under one identity:
\begin{align}
s_{i,e}^{\mathrm{init}}(a)
&=\mathbf{1}\!\left\{\begin{aligned}
&\exists u\in[0,a],\ \exists j\neq\noid:\\[-2pt]
&J_i(g_e,t)=j\quad
\forall t\in[\tau_e+u,\tau_e+u+1\,\mathrm{s}]
\end{aligned}\right\}, \label{eq:init_event}\\
S_i^{\mathrm{init}}(a)
&=\frac{\sum_{e\in\mathcal{E}_{\mathrm{init}}}s_{i,e}^{\mathrm{init}}(a)}
{|\mathcal{E}_{\mathrm{init}}|}.
\label{eq:init_rate}
\end{align}
Eq.~\eqref{eq:init_rate} therefore reports the cumulative fraction of trajectories stably established by delay $a$.

\section{Experimental Setup}\label{sec:setup}
\subsection{Data, detections, and evaluation}
We evaluate the 27-sequence JRDB test split in a common world frame using STEAM-LO odometry \cite{burnett_continuous-time_2025}. Every tracker receives the same post-NMS \personminkunet{} pedestrian detections, trained on the JRDB training split, with score $s\geq0.5$ \cite{jia_person-minkunet_2021}. No tracker uses a GPU.

\textbf{Evaluation correction.} We found that the official JRDB 3D-IoU routine used enclosing rather than intersection height for vertical overlap, inflating 3D IoU and HOTA. We corrected this and the routine's oriented-footprint clipping. All reported results use the corrected evaluator and are therefore not directly comparable to the official JRDB leaderboard.

\subsection{Compared trackers and runtime platform}
\begin{figure*}[!htb]
\centering
\includegraphics[width=0.92\textwidth]{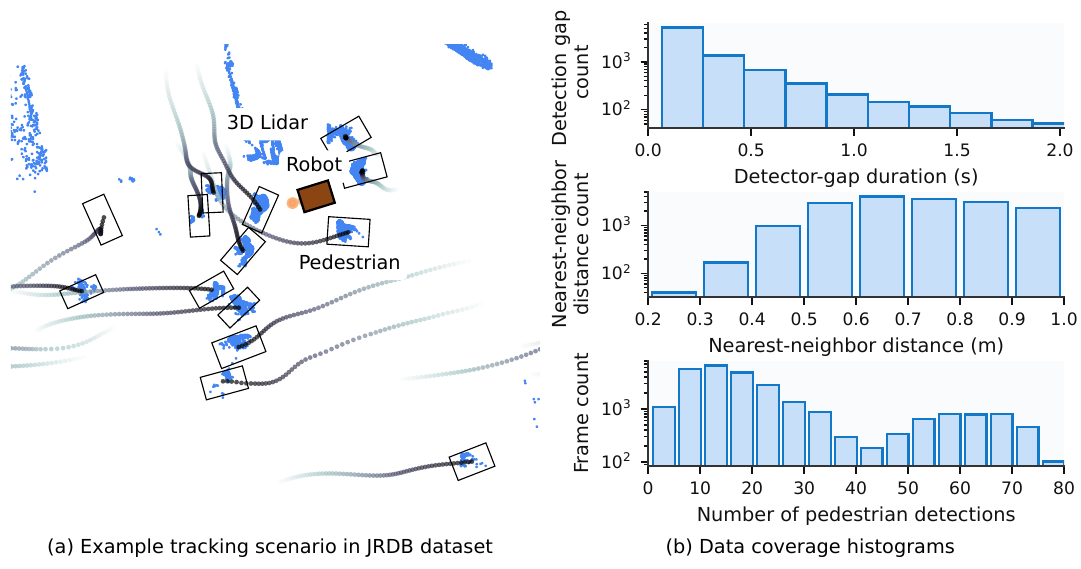}
\caption{\textbf{Pedestrian-centric setting and evaluation coverage.} (a) Example bird's-eye-view JRDB scene. (b) Counts supporting the detector-gap, nearest-neighbor, and runtime analyses.}
\label{fig:coverage}
\end{figure*}

We compare AB3DMOT \cite{weng_3d_2020}, Fast-Poly \cite{li_fast-poly_2024}, SimpleTrack \cite{pang_simpletrack_2022}, CBMOT \cite{benbarka_score_2021}, GNN-PMB \cite{liu_gnn-pmb_2023}, box-only ELPTNet \cite{guang_elptnet_2025}, PedRefTrack, and its GT-assisted variant. The available ELPTNet implementation did not fully reproduce the paper's architecture, so we enabled its documented output gate and bounded prediction through missed detections. Exposed parameters were tuned to maximize HOTA, with configurable output coasting set to the longest duration before HOTA decreased. Success-rate bands are 95\% bootstrap intervals from sequence resampling. Figure~\ref{fig:coverage}(b) reports sample counts for the gap, nearest-neighbor, and runtime strata.

Runtime is measured on an NVIDIA \orin{} 64\,GB in maximum-power mode. We use 10\,Hz as a conservative real-time target for common spinning-LiDAR frame rates.

\subsection{GT-derived detector perturbations}
\begin{figure}[!htb]
\centering
\includegraphics[width=0.46\textwidth]{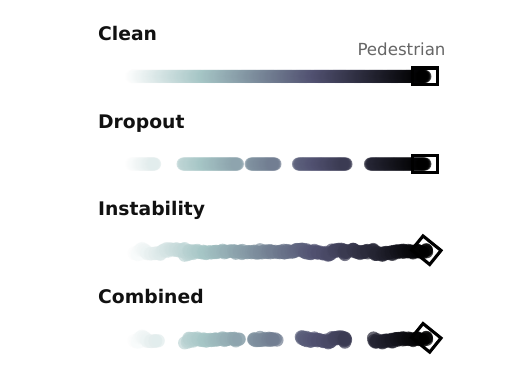}
\caption{\textbf{GT-derived detector-input models.} Dropout removes contiguous observations. Instability introduces perturbations to box center, orientation, and dimensions. Combined inputs apply matched dropout and instability severities.}
\label{fig:pseudo_modes}
\end{figure}

To examine how tracker HOTA performance changes as observation quality steadily degrades, we create pseudo detections from GT and apply controlled dropout and box-instability perturbations. Figure~\ref{fig:pseudo_modes} illustrates these failure modes. Dropout introduces increasingly long intervals without detector support, while instability perturbs box geometry through persistent trajectory-level biases and smaller frame-wise jitter. For instability, each trajectory draws $K$ persistent box-error hypotheses and probabilistically switches among them. The final comparison combines matched dropout and instability severities from Table~\ref{tab:pseudo_spec}, providing every tracker with the same controlled increase in input difficulty.

\begin{table}[!t]
\caption{\normalfont \textbf{Combined GT-derived perturbation severities.} Each track draws $K$ persistent box-error hypotheses and switches the active one with probability $p_{\mathrm{switch}}$. Bias terms $\sigma^{\mathrm{bias}}$ define hypothesis-specific fixed offsets from GT; jitter terms $\sigma^{\mathrm{jit}}$ define independent zero-mean noise added each frame. L/W/H values are relative.}
\label{tab:pseudo_spec}
\centering
\setlength{\tabcolsep}{3.5pt}
\renewcommand{\arraystretch}{1.07}
\begin{tabular}{@{}lccc@{}}
\toprule
Parameter & Mild & Moderate & Severe\\
\midrule
Dropout start $p_{\mathrm{drop}}$ (\%/frame) & $0.5$ & $1.0$ & $2.0$\\
Dropout duration (s) & $0.05$--$0.50$ & $0.05$--$1.00$ & $0.05$--$2.00$\\
Box-error hypotheses $K$ & $3$ & $4$ & $5$\\
Hypothesis switch $p_{\mathrm{switch}}$ (\%/frame) & $5$ & $10$ & $20$\\
XY $\sigma^{\mathrm{bias}}_{xy}/\sigma^{\mathrm{jit}}_{xy}$ (m)
    & $0.03/0.01$ & $0.06/0.02$ & $0.10/0.03$\\
Yaw $\sigma^{\mathrm{bias}}_{\psi}/\sigma^{\mathrm{jit}}_{\psi}$ (rad)
    & $0.50/0.10$ & $0.80/0.15$ & $1.10/0.20$\\
L/W/H $\sigma^{\mathrm{bias}}_{\mathrm{dim}}/\sigma^{\mathrm{jit}}_{\mathrm{dim}}$ (\%)
    & $3/1$ & $5/2$ & $8/3$\\
\bottomrule
\end{tabular}
\end{table}

\section{Results}\label{sec:results}
\subsection{Aggregate accuracy and deployment variation}
Table~\ref{tab:global_results} reports HOTA and its decomposition. GT-assisted PedRefTrack reaches 33.33\% HOTA, compared with 29.67\% for PedRefTrack and 27.48\% for Fast-Poly, corresponding to deficits of 3.66 and 5.85 percentage points. Methods without GT assistance span only 24.26--29.67\% HOTA, providing limited separation for tracker selection. LocA is narrower still at 65.49--66.15\%, indicating that the HOTA differences arise mainly from track-output and association behavior rather than localization.

\begin{figure*}[!htb]
\centering
\includegraphics[width=0.92\textwidth]{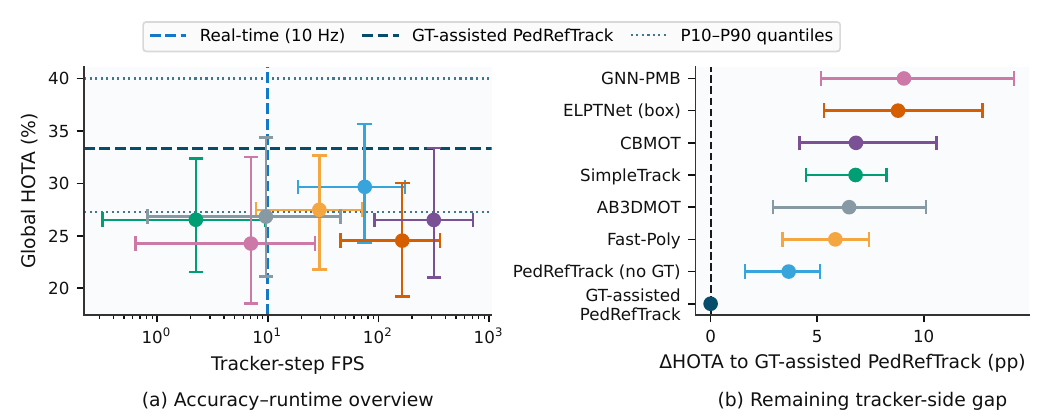}
\caption{Accuracy, runtime, and tracker-side headroom. (a) HOTA versus median tracker-step FPS. Whiskers show the 10th--90th percentile ranges of sequence HOTA (vertical) and frame-level FPS (horizontal). (b) Per-sequence HOTA deficit to GT-assisted PedRefTrack, with quantiles from paired same-sequence differences. GT-assisted PedRefTrack idealizes association and motion while retaining the remaining tracker pipeline.}
\label{fig:global_overview}
\end{figure*}

Figure~\ref{fig:global_overview} shows why a single split-wide number remains incomplete. The 10th--90th percentile ranges in panel (a) reveal large variation in both sequence HOTA and frame runtime. Panel (b) instead pairs every sequence with GT-assisted PedRefTrack before taking the difference. This removes the easy-versus-difficult sequence variation and shows the tracker-side deficit on the same scene. The varying ranges show that tracker choice affects both average accuracy and its consistency across scenes.

\begin{table*}[!t]
\vspace{1mm}
\caption{\normalfont \textbf{JRDB test results with shared detections.} $\dagger$ marks the GT-assisted variant. FPS measures tracker-step throughput on \orin{} with brackets showing the 10th--90th quantiles. CPU cores reports the number used out of 12 available. Bold and underlining mark the best value including and excluding the GT-assisted variant, respectively. No tracker uses a GPU. Results use our corrected JRDB 3D-IoU evaluator and are not directly comparable to the official JRDB leaderboard.}
\label{tab:global_results}
\centering
\setlength{\tabcolsep}{2.05pt}
\renewcommand{\arraystretch}{1.10}
\begin{tabular}{l|rrrrrrrrrcc}
\toprule
\multirow{2}{*}{Method} & \multirow{2}{*}{\shortstack{HOTA$\uparrow$\\(\%)}} &
\multicolumn{3}{c}{Detection (\%)} & \multicolumn{3}{c}{Association (\%)} &
\multirow{2}{*}{\shortstack{LocA$\uparrow$\\(\%)}} & \multicolumn{3}{c}{Runtime}\\
\cmidrule(lr){3-5}\cmidrule(lr){6-8}\cmidrule(lr){10-12}
& & DetA$\uparrow$ & DetRe$\uparrow$ & DetPr$\uparrow$ & AssA$\uparrow$ & AssRe$\uparrow$ & AssPr$\uparrow$ & & Med. FPS$\uparrow$ & \shortstack{P10--P90 FPS\\$\uparrow$} & \shortstack{CPU cores\\used$\downarrow$}\\
\midrule
GT-assisted PedRefTrack (ours)$\dagger$ & \textbf{33.33} & \textbf{29.34} & \textbf{36.16} & \textbf{43.20} & \textbf{38.62} & \textbf{42.54} & \textbf{57.51} & \textbf{66.26} & 62.26 & [17.47,150.13] & 1\\
\midrule
PedRefTrack (ours, no GT) & \underline{29.67} & \underline{27.18} & 33.79 & 41.28 & \underline{33.13} & \underline{37.80} & 51.43 & 65.74 & 74.96 & [18.71,172.39] & 1\\
Fast-Poly \cite{li_fast-poly_2024} & 27.48 & 25.61 & 32.23 & 39.83 & 30.27 & 33.85 & 52.52 & 66.12 & 29.15 & [7.77,70.55] & 3\\
AB3DMOT \cite{weng_3d_2020} & 26.83 & 26.84 & 33.65 & 40.99 & 27.72 & 31.05 & 52.93 & 65.95 & 9.59 & [0.82,45.20] & 7\\
SimpleTrack \cite{pang_simpletrack_2022} & 26.53 & 25.42 & \underline{34.34} & 36.52 & 28.52 & 32.10 & 52.13 & 65.49 & 2.23 & [0.32,9.54] & 7\\
CBMOT \cite{benbarka_score_2021} & 26.51 & 26.24 & 31.84 & 41.91 & 27.41 & 30.66 & 51.65 & 65.56 & \textbf{\underline{315.25}} & [91.45,714.42] & 2\\
ELPTNet (box) \cite{guang_elptnet_2025} & 24.53 & 25.27 & 30.07 & \underline{42.79} & 24.45 & 26.98 & \underline{53.09} & \underline{66.15} & 163.52 & [45.49,358.56] & 7\\
GNN-PMB \cite{liu_gnn-pmb_2023} & 24.26 & 25.95 & 31.43 & 42.00 & 23.36 & 26.12 & 50.21 & 65.64 & 7.02 & [0.64,26.76] & 2\\
\bottomrule
\end{tabular}
\end{table*}

\subsection{Trajectory tests expose different failure mechanisms}
\begin{figure*}[!htb]
\centering
\includegraphics[width=0.92\textwidth]{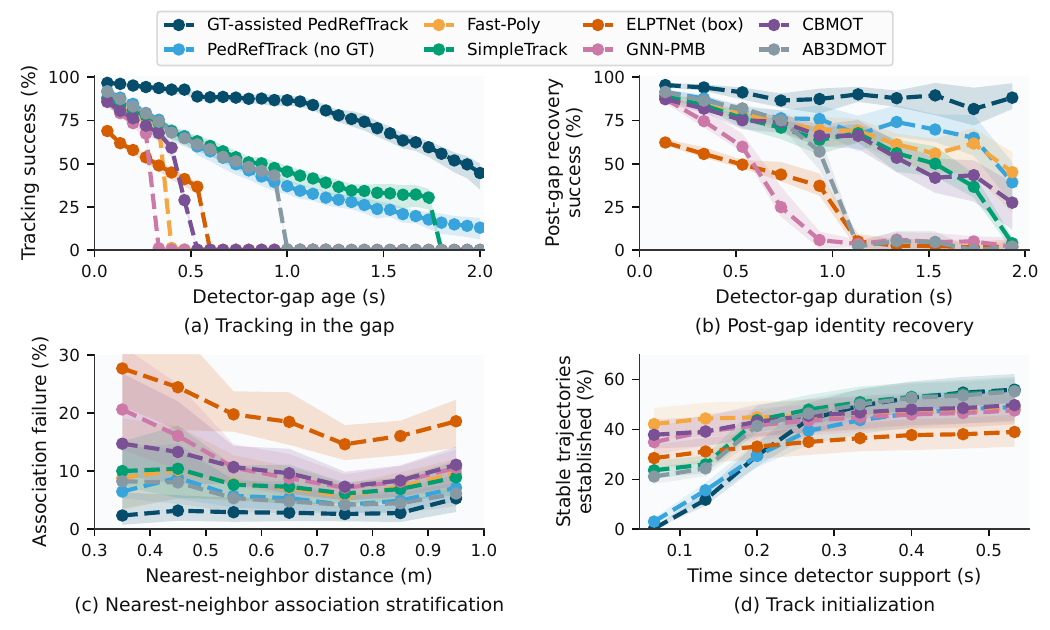}
\caption{\textbf{Trajectory-level success tests on fixed detections.} (a) Same-ID, spatially accepted output through detector gaps. (b) Pre-gap identity recovery when detector support returns, regardless of gap output. (c) Same-ID continuity failure over $1.0$\,s windows versus 10th-percentile GT nearest-neighbor distance. (d) Cumulative stable initialization within $0.5$\,s, requiring a subsequent second of uninterrupted correct output. Shading shows 95\% bootstrap intervals from sequence resampling.}
\label{fig:capabilities}
\end{figure*}

The shared-input tests in Fig.~\ref{fig:capabilities} reveal differences hidden by aggregate results. In panel (a), GT-assisted PedRefTrack retains the original, spatially accepted identity through $1.0$\,s in $86.8\%$ of eligible gaps. SimpleTrack reaches $45.4\%$ and PedRefTrack $37.0\%$, while the remaining methods are below $1\%$. Thus, a tracker may continue publishing yet retain spatially correct, same-ID output in fewer than half the gaps. The decline is consistent with open-loop extrapolation becoming unreliable over longer gaps, while lifecycle policy also limits published prediction duration. Gap success therefore reflects both motion prediction and output-lifetime policy. Identity recovery is markedly stronger than published continuation. At $0.5$\,s, Fast-Poly, GNN-PMB, and CBMOT maintain valid output through about $0.2\%$ of gaps, yet recover pre-gap identity in $78.9\%$, $59.8\%$, and $75.1\%$. At $1.5$\,s, PedRefTrack, Fast-Poly, SimpleTrack, and CBMOT recover $69.8\%$, $55.8\%$, $50.0\%$, and $41.9\%$, while AB3DMOT, GNN-PMB, and ELPTNet remain below $5\%$. The recovery contrast reflects lifecycle differences, as some methods discard identity state when coasting ends while others retain dormant state for re-association.

Figure~\ref{fig:capabilities}(c) shows continuity worsening with pedestrian-pedestrian distance. As nearest-neighbor distance decreases from $0.7$--$0.8$\,m to $0.3$--$0.4$\,m, failure rises across trackers without GT assistance, by $2.3$ percentage points for PedRefTrack and up to $13.5$ for GNN-PMB. The deficit to GT-assisted PedRefTrack also widens, from $2.0$--$15.7$ at $0.6$--$0.7$\,m to $4.1$--$25.2$ in the closest bin. Close-neighbor association failures remain smaller than during $1.0$\,s detector gaps, making gap handling and motion extrapolation the larger limitation in this JRDB case study, while proximity becomes most consequential below about $0.5$\,m.

Initialization also varies substantially across trackers. By $0.5$\,s, $39.1$--$56.0\%$ of shared trajectory starts are stably established, with larger differences at shorter delays. At $0.2$\,s, Fast-Poly has established $45.0\%$ of starts versus $29.4\%$ for PedRefTrack, showing substantially earlier establishment. By $0.4$\,s, PedRefTrack reaches $46.5\%$ versus $46.8\%$, with higher detection precision ($41.3\%$ versus $39.8\%$) and much longer valid coasting in panel (a). These results expose an initialization trade-off: earlier output reduces establishment delay, whereas more selective confirmation can improve detection precision at the cost of later availability.

\subsection{Runtime depends on detection load}
\begin{figure}[!htb]
\centering
\includegraphics[width=0.46\textwidth]{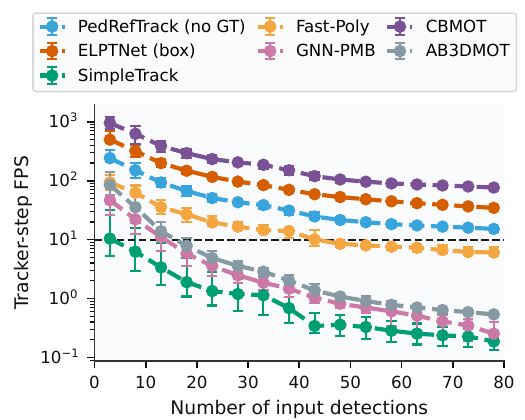}
\caption{\textbf{Tracker-step FPS versus input detections.} Whiskers show the 10th--90th quantiles, and the dashed line marks the 10\,Hz LiDAR rate. Detector inference is excluded.}
\label{fig:runtime_load}
\end{figure}

Tracker-step throughput decreases strongly with detection count (Fig.~\ref{fig:runtime_load}), with several methods falling below the 10\,Hz target as load increases. CBMOT, ELPTNet, and PedRefTrack remain above 10 Hz at the highest displayed loads, with PedRefTrack retaining about 15.2 FPS in the 75--80 detection bin. Fast-Poly falls below 10\,Hz at roughly 45 detections, while AB3DMOT, GNN-PMB, and SimpleTrack do so at substantially lower loads.

\begin{figure}[!htb]
\centering
\includegraphics[width=0.46\textwidth]{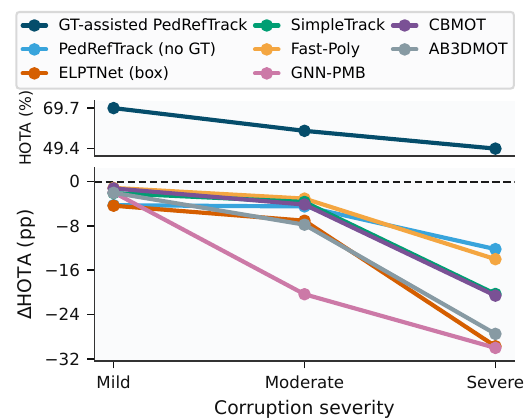}
\caption{GT-derived detector perturbations. The upper strip shows GT-assisted PedRefTrack HOTA. The main panel shows each tracker's HOTA deficit to this reference. Mild, moderate, and severe correspond to Table~\ref{tab:pseudo_spec}.}
\label{fig:pseudo_combined}
\end{figure}

\subsection{Controlled GT-derived perturbations}
The main panel in Fig.~\ref{fig:pseudo_combined} shows each tracker's HOTA deficit to GT-assisted PedRefTrack as dropout and box instability increase from mild to severe. The deficits are compressed under mild corruption but separate substantially with increasing severity. PedRefTrack and Fast-Poly show the smallest deficit growth, indicating greater robustness to the modeled observation loss and instability, while the remaining trackers lose progressively more ground to the GT-assisted reference. Tracker choice therefore becomes more consequential as observation support degrades. Similar aggregate robustness does not imply similar trajectory behavior: PedRefTrack and Fast-Poly remain close across severities, yet Fig.~\ref{fig:capabilities} shows markedly different initialization, gap-continuation, and identity-recovery profiles.

\section{Discussion and Limitations}\label{sec:discussion}

While scalar benchmark values like HOTA provide a useful, compact aggregate summary, they do not directly reveal which deployment-relevant tracker behaviors are reliable or where failures occur. Our protocol complements this view with intuitive success rates for defined tracking situations and runtime measured across input loads, while shared detections keep the comparison focused on tracker behavior. Applied to JRDB, the protocol shows that methods with similar HOTA reach that aggregate performance through different trade-offs. Some establish trajectories earlier, others maintain valid output longer through missed observations or retain identity for later recovery, and their runtime scales differently with pedestrian load, which is critical for robots with limited onboard compute. By exposing these properties separately, the protocol provides an interpretable profile of tracker behavior and identifies which capabilities remain limiting in a given pedestrian-centric environment.

The capability profiles suggest tracker-design priorities. In JRDB, detector-gap continuation degrades more severely than close-neighbor association, pointing to motion prediction during detection gaps as the largest limitation among tested properties. Gap success depends on both motion extrapolation and lifecycle policy, but its sharp decline with gap duration is consistent with open-loop prediction becoming unreliable when pedestrian motion changes. This motivates richer pedestrian-motion models incorporating local interactions and scene context. The proximity and recovery results motivate pedestrian-specific structural and temporal features beyond box geometry, which could disambiguate nearby pedestrians and support post-gap re-association when return positions deviate from motion predictions. Several trackers recover identity far more often than they maintain valid gap output. Together with initialization results, this motivates separating confirmation, published coasting, and hidden identity retention so that uncertain positions can be suppressed without discarding state useful for later re-association.

The study is limited to one JRDB-trained detector operating point and one pedestrian-centric dataset. GT-derived perturbations broaden the input conditions but do not reproduce the full error structure of a learned detector, and the reported profiles reflect HOTA-tuned configurations under selected matching thresholds and temporal horizons rather than each tracker's full operating range. Runtime covers only tracker-step execution on one embedded platform and excludes the remaining perception and navigation stack. Finally, the tests are deployment-oriented proxies rather than direct navigation measures. These limitations motivate validation across further datasets, detector regimes, tracker settings, hardware platforms, and downstream navigation tasks.

\section{Conclusion}\label{sec:conclusion}
We presented a deployment-focused, tracker-only protocol that complements HOTA with directly interpretable success profiles and load-aware embedded runtime evaluation under shared detections. Applied to JRDB, the protocol shows that trackers with similar HOTA can differ substantially in initialization, detector-gap continuation, identity recovery, close-neighbor behavior, and runtime scaling. PedRefTrack remains competitive across aggregate accuracy and capability profiles while exceeding 10\,Hz at the highest tested loads on a single \orin{} CPU core. Among the tested properties, detector-gap continuation is the largest limitation, while post-gap identity recovery remains much stronger and close-neighbor failures increase mainly at short separations. These results support complementing aggregate measures such as HOTA with deployment-relevant capability profiles to better assess tracker suitability in a given environment.

\balance
\bibliographystyle{IEEEtran}
\bibliography{IEEEabrv,RAL_tracking,bstcontrol}

\end{document}